# CUID: A NEW STUDY OF PERCEIVED IMAGE QUALITY AND ITS SUBJECTIVE ASSESSMENT

*Lucie Lévêque[1], Ji Yang[2], Xiaohan Yang[3], Pengfei Guo[4], Kenneth Dasalla[5], Leida Li[6], Yingying Wu[7], and Hantao Liu[2]*

[1]Laboratory Ergonomics and Cognitive Sciences applied to Transport, Gustave Eiffel University, Lyon, France
[2]School of Computer Science and Informatics, Cardiff University, Cardiff, United Kingdom
[3]School of Electronic and Information Engineering, Xi'an Jiaotong University, Xi'an, China
[4]School of Computational Science, Zhongkai University of Agriculture and Engineering, Guangzhou, China
[5]Department of Computer Science, University of Bath, Bath, United Kingdom
[6]School of Artificial Intelligence, Xidian University, Xi'an, China
[7]School of Mathematics, Cardiff University, Cardiff, United Kingdom

## ABSTRACT

Research on image quality assessment (IQA) remains limited mainly due to our incomplete knowledge about human visual perception. Existing IQA algorithms have been designed or trained with insufficient subjective data with a small degree of stimulus variability. This has led to challenges for those algorithms to handle complexity and diversity of real-world digital content. Perceptual evidence from human subjects serves as a grounding for the development of advanced IQA algorithms. It is thus critical to acquire reliable subjective data with controlled perception experiments that faithfully reflect human behavioural responses to distortions in visual signals. In this paper, we present a new study of image quality perception where subjective ratings were collected in a controlled lab environment. We investigate how quality perception is affected by a combination of different categories of images and different types and levels of distortions. The database will be made publicly available to facilitate calibration and validation of IQA algorithms.

## 1. INTRODUCTION

Image quality assessment (IQA) forms the basis of algorithms for the evaluation, monitoring, and optimisation of modern digital imaging systems. Over the last two decades, substantial progress has been made in the development of IQA models that can automatically quantify image quality as perceived by humans. The fundamental research underpinning this field encompasses two important elements: first, visual perception experiments that provide perceptual evidence and subjective data for model calibration and validation; second, computational algorithms that give objective predictions of perceived quality. Many IQA models have been created to successfully capture image quality aspects in a variety of application domains [1]-[2].

However, IQA models have been intrinsically designed or trained with inadequate subjective data with a small degree of stimulus variability due to practical difficulties in conducting reliable larger-scale subjective experiments. This has limited the capability of these IQA models to generalise in real-world circumstances where image content is diverse and complex [3]-[4]. The commonly used subjective databases include LIVE [5], CSIQ [6], IVC [7], TID [8] and others as listed in [9]. The subjective data have made a significant contribution to the development and benchmarking of objective IQA models; however, it is critical to be aware of the limitations of subjective studies and the associated implications for the performance of IQA models. The two essential limitations are: 1) a small degree of stimulus variability in terms of the amount and diversity of scene content, covering a tiny portion of the space of digital images; and 2) bias/noise of the subjective data due to the uncontrolled variables contained in an experiment, such as inconsistent viewing conditions and biased scoring protocol. Both limitations have the potential to negatively impact the design and evaluation of IQA algorithms.

Very little research has been done to investigate image quality perception with a larger degree of stimulus variability and under fully controlled experimental conditions. In this study, we aim to understand how perceived quality is affected by a combination of various categories of digital content and different types and levels of distortions. More importantly, we adopt a within-subjects experimental design [10], in which all participants perceive and score the entire set of stimuli. This gives the largest dataset of its kind, namely **C**ardiff **U**niversity **I**mage quality **D**atabase (CUID).

## 2. MATERIAL AND METHODS

### 2.1. Stimuli

The source images used to construct our new database were sixty high-resolution and high-quality images collected from the Unsplash website [11]. These images were all resized (using bicubic interpolation) to 1920×1080 pixels. To study the impact of image content in a more systematic way, the images were selected and assigned into the following ten categories. Action (ACT): images that show high activity, Black and White (BNW): grayscale images, Computer-Generated Imagery (CGI): computer-generated graphic images, Indoor (IND): images captured from the indoor scenarios, Object (OBJ): images of various objects, Outdoor Man-made (ODM): images captured from outdoor scenarios with man-made objects, Outdoor Natural (ODN): images captured from outdoor scenarios with nature scenes, Pattern (PAT): images with a set of repeating objects, Portrait (POT): close-up shots of human faces, and Social (SOC): images with interactions between people. Figure 1 shows the source images and categories.

The source images were distorted using three different image distortion types occurring in real-world applications: contrast change (CC), JPEG compression (JPEG), and motion blur (MB). These distortions reflect three distinctive image impairments, i.e., CC affects the colours of images, JPEG yields local artifacts, whereas MB causes a global distortion. The contrast change, JPEG compression and motion blur were implemented by MATLAB's *adapthisteq* function, *fspecial* function, and *imwrite* function, respectively. As illustrated in Figure 2, for each distortion type, three distorted versions (i.e., Q1, Q2 and Q3) per source image were generated by varying the distortion parameters. Visual inspection by image quality experts (mainly the authors) was carried out during the database construction process to ensure that the distorted images reflect three distinctive levels of perceived quality: Q1 (i.e., with perceptible but not annoying artifacts), Q2 (i.e., with noticeable and annoying artifacts) and Q3 (i.e., with very annoying artifacts). Therefore, the source images were distorted by three distortion types and at three distortion levels, resulting in **600** stimuli (including the originals).

It should be noted that the number of source images and the number of distortion types/levels are chosen in our study to make a balanced database that contains adequate diversity in image content and quality variations (i.e., degree of stimuli variability) while allowing within-subjects (i.e., reliability of subjective scores) scoring practically feasible. Also, it is worth noting that a good IQA database should have a uniform distribution of stimuli in the perceptual quality range. A highly non-uniform distribution of subjective scores would potentially bias the calibration and validation of objective IQA algorithms. Figure 3 shows the histogram of the subjective scores for the entire database (after score processing as detailed in Section 3). It can be seen that the stimuli are fairly evenly distributed in the quality range, and the distribution is similar to that of the LIVE database [5].

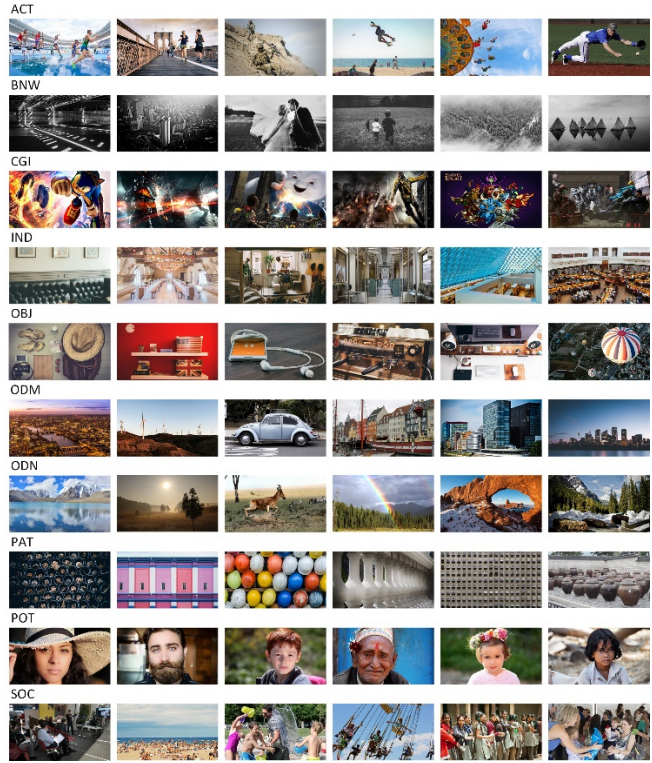

Fig. 1. Illustration of the sixty source images used in our study.

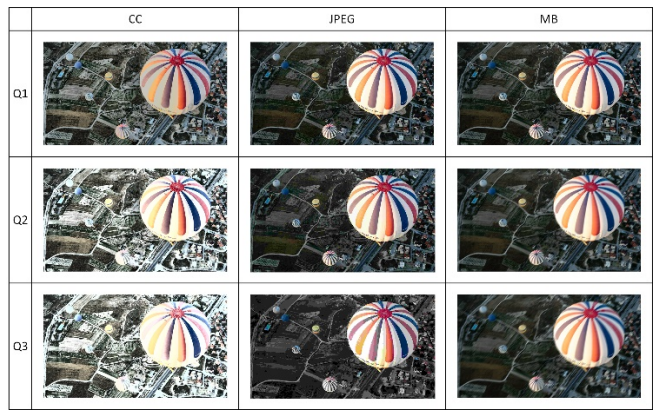

Fig. 2. Illustration of sample distorted images used in our study.

## 2.2. Experimental procedure

We conducted a visual perception experiment in the Visual Computing laboratory at Cardiff University in a standard office environment [12]. The laboratory represented a fully controlled viewing environment to ensure consistent experimental conditions, i.e., low surface reflectance and approximately constant ambient light. The test stimuli were displayed on a 19-inch LCD screen, with a native resolution of 1920×1080 pixels. The viewing distance was maintained around 60cm. A standardised single-stimulus method [12] was used, where subjects were each asked to score the quality of all the 600 stimuli contained in the database. The scoring scale ranged from 0 to 100 and was divided into five portions with the following semantic labels: Bad: [0-20], Poor: [20-40], Fair: [40-60], Good: [60-80], and Excellent: [80-100]. The within-subjects method, in which each subject views all stimuli, can produce reliable subjective ratings, but is prone to carry-over effects such as fatigue. To reduce the undesirable effects, we divided the database into four partitions of 150 stimuli each; thus, each subject had to complete four sessions with a "break" period of one day between sessions. For each subject, the stimuli were first randomised before partitioning, and the stimuli in each session were presented to each subject in a random order. Nineteen participants naïve to image quality assessment, including eight males and eleven females and in the 23-52 age range, participated in the experiment. Before the start of the first experimental session, each participant was provided with written instructions on the procedure of the experiment. A training session was conducted to familiarise the participants with the stimuli and distortions involved, and with how to use the range of the scoring scale. The images used in the training session were different from those used in the real experiment.

## 3. EXPERIMENTAL RESULTS

### 3.1. Processing of raw data

First, *z*-scores were calculated to account for the differences between subjects in the use of the scoring scale and calibrate them towards the same mean and standard deviation. The raw subjective scores were converted into *z*-scores as follows:

$$z_{ij} = (r_{ij} - \mu_i)/\sigma_i \quad (1)$$

where $r_{ij}$ denotes the raw score given by the *i*-th subject to the *j*-th test stimuli, $\mu_i$ is the mean of all scores for the subject *i*, and $\sigma_i$ is the corresponding standard deviation. Second, an outlier detection and subject removal procedure as suggested in [13] was performed, resulting in 4% of scores being detected and removed as outliers, and no participant being rejected. Finally, the mean opinion score (MOS) of each stimulus was computed as the mean of the remaining *z*-scores over all subjects:

$$MOS_j = \frac{1}{s}\sum_{i=1}^{s} z_{ij} \quad (2)$$

where *s* is the number of remaining subjects for the *j*-th image. To make the final scores easier to interpret, the resulting mean opinion scores were linearly remapped to the range of [0, 100], as the histogram shown in Figure 3.

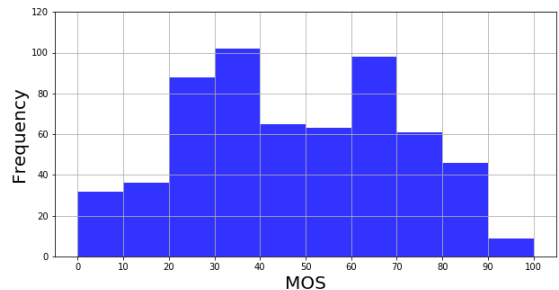

Fig. 3. Illustration of the histogram of subjective ratings (MOS) of stimuli contained in the CUID database.

### 3.2. Properties of CUID database

To evaluate the validity of the resulting MOS, we quantify the variation in scoring between human subjects, using the correlation, i.e., the Spearman Rank Correlation Coefficient (SRCC) between the MOS and each subject's scores. Figure 4 shows the SRCC values for all subjects, as well as the mean SRCC value. It can be seen that there is a high agreement between subjects in scoring the quality of test stimuli.

Now, one of the unique features of the CUID database is that the impact of categorical variables, such as image content classification on quality, can be statistically analysed. This is due to the advantage of the use of within-subjects method, where the differences between MOS values contained in the entire database are meaningful [14]. It should be noted that MOS values obtained from multiple sessions (of different sets of images) using different subjects cannot be combined for statistical analysis [14] unless a sophisticated scale-realignment experiment [5] is performed. Figure 5 illustrates the category-wise MOS for the CUID database. It can be seen from the figure that the mean opinion score of the CGI and OBJ content is higher than that of other categories of image content. This suggests that CGI and OBJ content seems to be (relatively) less affected by the distortion. The SOC content was given the lowest mean opinion score in quality across all categories. There is some evidence to indicate that the category of image content (potentially linked to more sophisticated cognitive processes, such as emotion) tends to affect image quality perception even when the same distortion is applied (note the same distortion types and levels were equally implemented for each category of CUID database).

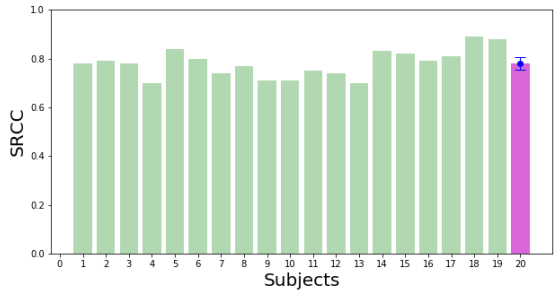

Fig. 4. Illustration of the correlation (SRCC) between MOS and individual subjective scores. The right-most bar shows the mean correlation with a 95% confidence interval.

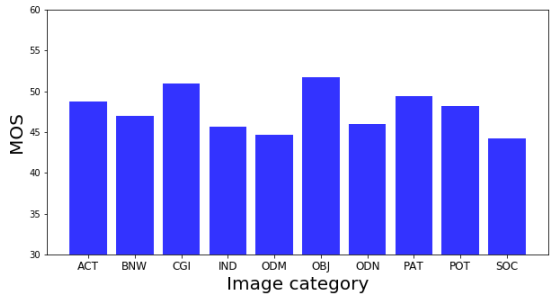

Fig. 5. Category-wise MOS of the CUID database.

### 3.3. Preliminary test of IQA metrics

Preliminarily, we test the performance of popular image quality assessment (IQA) metrics on the CUID database. The test is limited to three full-reference IQA metrics: PSNR [15], SSIM [16], and VIF [17]. These metrics are chosen because they give consistent performance while the need for calibrating model parameters is minimum. Figure 6 shows the scatter plots of IQA metrics. The performance of IQA metrics is evaluated against CUID database by using Pearson Linear Correlation Coefficient (PLCC), and Spearman Rank Order Correlation Coefficient (SROCC) [14]. Table 1 shows the results of the performance evaluation. Note following the approach taken in [13], results are reported without nonlinear fitting in order to better visualise differences in IQA performance. In general, these IQA metrics' performance is unsatisfactory as it shows low PLCC and SROCC values. This suggests that there is still room for improvement on IQA metrics in terms of handling diverse and complex image content.

Table 1. Performance of image quality assessment (IQA) metrics, based on the CUID database.

| IQA metric | PLCC | SROCC |
|---|---|---|
| PSNR | 0.12 | 0.14 |
| SSIM | 0.51 | 0.54 |
| VIF | 0.59 | 0.69 |

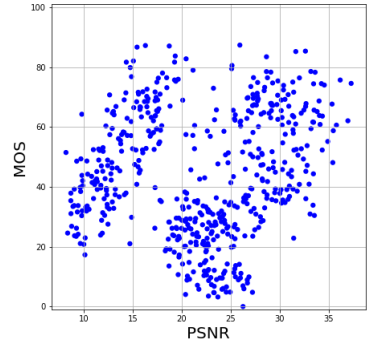

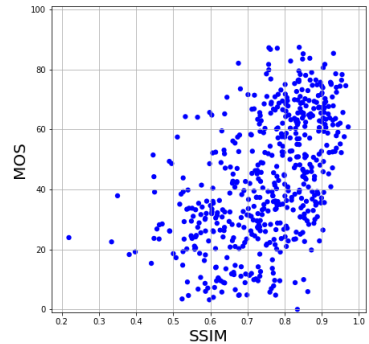

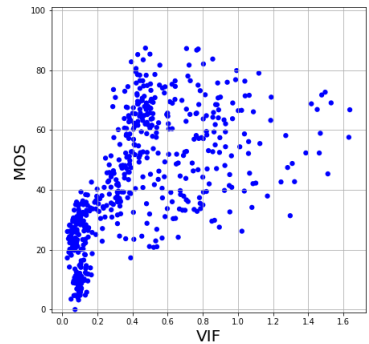

Fig. 6. Scatter plots of MOS versus image quality assessment (IQA) metrics (PSNR, SSIM, VIF), using the CUID database.

### 4. CONCLUSIONS

In this paper, we presented a new image quality assessment database, namely CUID. This database is best of its kind in terms of being constructed under a fully controlled laboratory environment, using a reliable within-subjects scoring method while having a large degree of stimulus content variability (i.e., 60 source images of 10 distinctive categories, giving a total of 600 test stimuli). The CUID database poses some new challenges to the image quality research community, including the impact of image classification (category) on perceived quality, and the improvement of IQA metrics' ability to handle diverse and complex image content.